# Automating Customer Service using LangChain

Building custom open-source GPT Chatbot for organizations

**Keivalya Pandya**
19me439@bvmengineering.ac.in
Birla Vishvakarma Mahavidyalaya, Gujarat, India

**Prof. Dr. Mehfuza Holia**
msholia@bvmengineering.ac.in
Birla Vishvakarma Mahavidyalaya, Gujarat, India

*Abstract*— In the digital age, the dynamics of customer service are evolving, driven by technological advancements and the integration of Large Language Models (LLMs). This research paper introduces a groundbreaking approach to automating customer service using LangChain, a custom LLM tailored for organizations. The paper explores the obsolescence of traditional customer support techniques, particularly Frequently Asked Questions (FAQs), and proposes a paradigm shift towards responsive, context-aware, and personalized customer interactions. The heart of this innovation lies in the fusion of open-source methodologies, web scraping, fine-tuning, and the seamless integration of LangChain into customer service platforms. This open-source state-of-the-art framework, presented as "Sahaay," demonstrates the ability to scale across industries and organizations, offering real-time support and query resolution. Key elements of this research encompass data collection via web scraping, the role of embeddings, the utilization of Google's Flan T5 XXL, Base and Small language models for knowledge retrieval, and the integration of the chatbot into customer service platforms. The results section provides insights into their performance and use cases, here particularly within an educational institution. This research heralds a new era in customer service, where technology is harnessed to create efficient, personalized, and responsive interactions. Sahaay, powered by LangChain, redefines the customer-company relationship, elevating customer retention, value extraction, and brand image. As organizations embrace LLMs, customer service becomes a dynamic and customer-centric ecosystem.

## I. Introduction

"Customer is king" is the ancient mantra reflecting the significance of customers in every business. In the digital age, where the rhythms of modern life are guided by the pulse of technology, the realm of customer service stands as the frontline of engagement between businesses and their clientele. It is the place where queries are answered, problems are resolved, and trust is forged.

This research paper brings the future of customer service, where automation, personalization, and responsiveness converge to redefine the customer-company relationship. At the heart of this transformation lies the integration of LLMs, exemplified by LangChain [1].

In the annals of customer service history, FAQs and traditional support mechanisms have long held sway. These venerable tools have dutifully served as repositories of information, attempting to address the queries and concerns of customers. However, as we stand at the cusp of a new era in customer service automation, it becomes abundantly clear that the traditional methods once hailed as revolutionary, are gradually becoming obsolete.

This paper is an invitation to envision a future where customer service is not a cost center but a wellspring of customer satisfaction and loyalty. We propose an open-source framework that can be scaled to any industry or organization to fulfill the consumer needs for support and query resolution within seconds.

For demonstration purposes, we use the information presented by Birla Vishvakarma Mahavidyalaya (BVM) Engineering College on their website *https://bvmengineering.ac.in/* as the context for our chatbot, from where it can retrieve all the information in real-time and answer to any queries that are raised by the users. Here, users can be anyone ranging from prospective students, current students who intend to get information from the Notice Board, researchers who wish to search for their potential research guide, and so on. The applications are endless.

## II. Literature Survey

S. Kim (2023) et al addresses the challenge of deploying resource-intensive large neural models, such as Transformers, for information retrieval (IR) while maintaining efficiency. Experimental results on MSMARCO benchmarks demonstrate the effectiveness of this approach, achieving successful distillation of both dual-encoder and cross-encoder teacher models into smaller, 1/10th size asymmetric students while retaining 95-97% of the teacher's performance [2]. L. Bonifacio et al (2022) highlights the recent transformation in the Information Retrieval (IR) field, propelled by the emergence of large pretrained transformer models. The MS MARCO dataset played a pivotal role in this revolution, enabling zero-shot transfer learning across various tasks [3].

This paper proposed a novel open-source approach to building LLM Chatbots using custom knowledge from the content in the website. It is unique in several ways:

1. We propose an open-source framework which is robust with the type of dataset available on the webpage or the web of links.

2. This implementation aims to compliment the use of FAQs with a more interactive and user-friendly interface.

3. We then do a comparative study of various models, their performance on the provided data relative to the expected response from the LLM.

## III. METHODOLOGY

This section covers the data collection, details about the selected model, fine-tuning, and integration with the Gradio APIs for web deployment.

### A. Data Collection

To gather the necessary data for our project, we employed BeautifulSoup web scraping techniques to retrieve publicly accessible information from an organization's homepage. We observed this page is often linked with all the relevant information required for the user/visitor. This approach allowed us to collect a wide array of data, including customer service FAQs, product manuals, support forums, chat logs, associated institutions, and so on. This data further serves as the context for our LLM.

### B. Embeddings

Embeddings play a pivotal role in the development of any LLM powered. They are vector representations of words or phrases in a continuous mathematical space that capture semantic and contextual information, allowing the model to understand the meaning and relationships between words, which is essential for providing meaningful responses to user queries.

We have used HuggingFace Instuct Embeddings – "hkunlp/instructor-large" a text embedding model fine-tuned for specific tasks and domains, such as classification, retrieval, clustering, and text evaluation [4]. What sets Instructor apart is its ability to generate tailored text embeddings without requiring additional fine-tuning. These embeddings are then stored using FAISS (Facebook AI Similarity Search) library that allows developers to quickly search for embeddings of multimedia documents that are similar to each other [5].

### C. Language Model

We have chosen Google's Flan T5 XXL as the most appropriate language model after comparing with other Flan T5 distributions to retrieve knowledge from the *vectorspace* and *chat_history* (or memory) [6]. The model retains the context of previous messages and uses that as a reference to predict answers for the upcoming questions. This helps users to have an interactive conversation with the chatbot, instead of a monotonous and robotic one.

### D. Integration with Customer Service Platforms

A simple chat window can be activated at the corner of any website which would enable users to interact with the chatbot and ask any relevant questions or doubts regarding the organization. However, for the demonstration purpose of this paper, we are using Gradio API framework [7].

## IV. RESULTS

In this section, we mention the metrics of comparison, provide comparative analysis, and use cases in association with an educational institution.

### A. Evaluating the Performance of LLMs

It is relatively difficult to evaluate LangChain agents, especially when trained on large chunks of context datasets for information retrieval. Hence, the current solution for the lack of metrics is to rely on human knowledge to get a sense of how the chain/agent is performing.

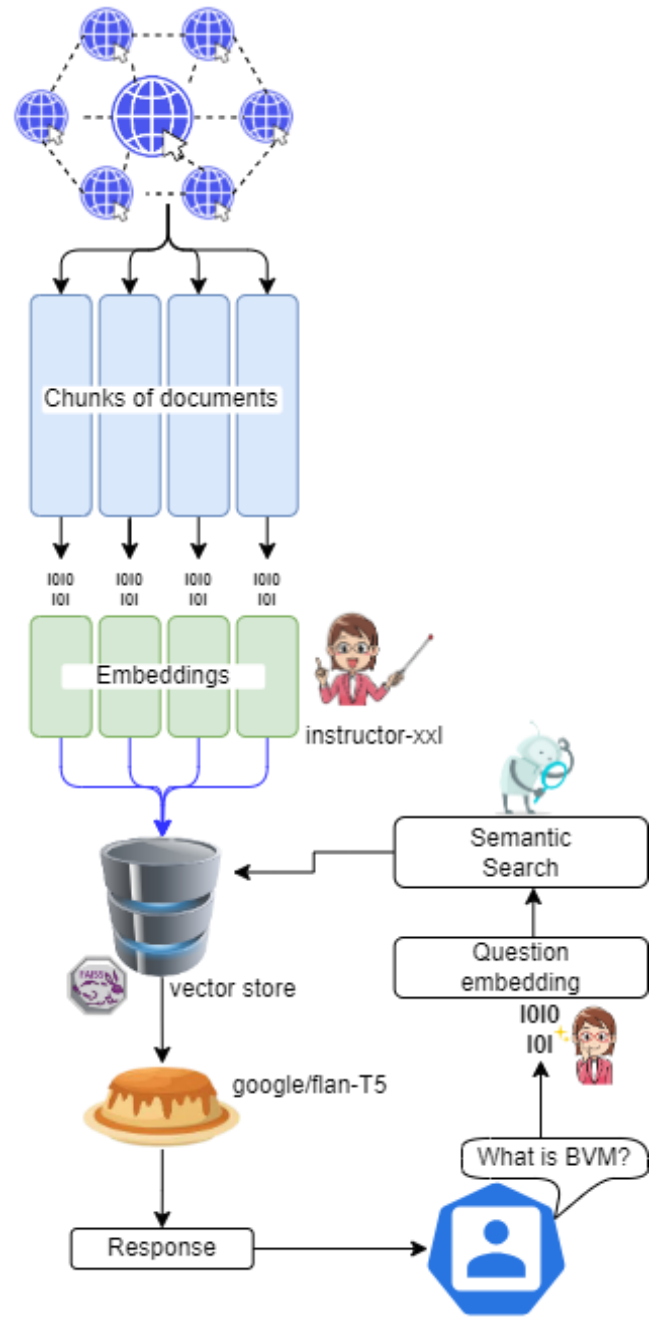

Fig. 1. Model Architecture – Sahaay

It is evident from TABLE-I, II and III that the XXL model outperforms other competitive LLMs such as BASE and SMALL.

TABLE I. Google's Flan-T5-*XXL* Performance

| Sr. No. | Query/Prompt | Answer | Performance |
|---|---|---|---|
| 1. | What is BVM? | Birla Vishvakarma Mahavidyalaya | ★★★★ |
| 2. | Where is it?[a] | Vallabh Vidyanagar, Gujarat India | ★★★ |
| 3. | What is IEEE BVM? | Institute of Electrical and Electronics Engineers Student Branch of BVM | ★★★★ |
| 4. | What is TRS BVM? | BVM Student Chapter associated with Robotics Society India | ★★★★ |
| 5. | Name the annual newsletter? | Vishvakarma Magazine and Newsletter | ★★★★ |
| 6. | Tell me something about ICWSTCSC | The International Conference for Women in Science and Technology Creating Sustainable Career is the 3rd International Conference happening in hybrid mode. | ★★★★★ |

[a.] Follow-up question

TABLE II. GOOGLE'S FLAN-T5-*BASE* PERFORMANCE

| Sr. No. | Query/Prompt | Answer | Performance |
|---|---|---|---|
| 1. | What is BVM? | BVM is the first engineering college of Gujarat State established way back in 1948. | ★★★★ |
| 2. | Where is it?[b] | VV Nagar, Gujarat, India | ★★★ |
| 3. | What is IEEE BVM? | first engineering college of Gujarat State established way back in 1948 | ★ |
| 4. | What is TRS BVM? | CVM BVM Robotics Society Of India | ★★ |
| 5. | Name the annual newsletter? | Vishvakarma Magazine | ★★★★ |
| 6. | Tell me something about ICWSTCSC | BVM 3rd Internartional Conference | ★★★★ |

[b.] Follow-up question

TABLE III. GOOGLE'S FLAN-T5-*SMALL* PERFORMANCE

| Sr. No. | Query/Prompt | Answer | Performance |
|---|---|---|---|
| 1. | What is BVM? | BVM is the first Autonomous Engineering institute of Gujarat to obtain academic autonomy for all its UG & PG programs, from University Grant Commission (UGC). | ★★★ |
| 2. | Where is it?[c] | BVM Engineering College | ★ |
| 3. | What is IEEE BVM? | engineering college of Gujarat State | ★ |
| 4. | What is TRS BVM? | BVM | ★ |
| 5. | Name the annual newsletter? | Campus Publications | ★ |
| 6. | Tell me something about ICWSTCSC | ICWSTCSC 2023 PMSSS students' reporting after admission (AY: 2023-24) at BVM 3rd Internartional Conference. | ★★★ |

[c.] Follow-up question

## B. Applications

Customer service establishes a direct connection between the customer and the company. It retains customers and extracts higher value from them. By harnessing the power of Large Language Models as shown in Fig. 2, customer service can be elevated to new heights, facilitating efficient, personalized, and responsive interactions. The LangChain fine-tuned over custom knowledge of the product, service, or organization can effectively address a wide array of customer inquiries and issues. Its ability to understand context and history empowers it to provide personalized support to customers. Automated customer service powered by LLMs is available around the clock and is also proficient in multiple languages.

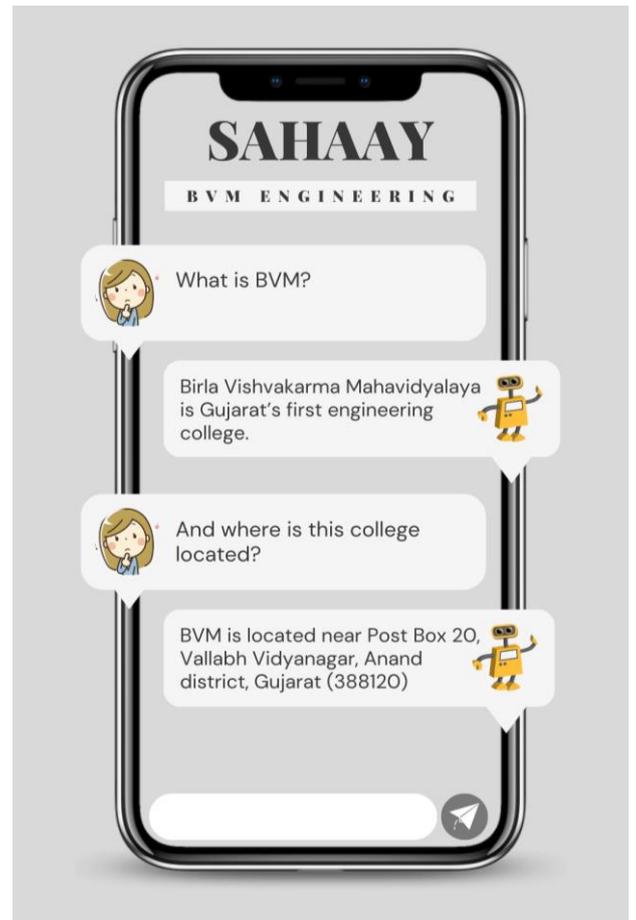

Fig. 2. User interface – Gradio framework

## V. CONCLUSION

In the ever-evolving landscape of customer service, the introduction of Sahaay's innovative approach presented in this paper, using LangChain as a prime example, ushered in a new era of automation. Automating customer service using Sahaay's open-source Large Language architecture leveraging LangChain revolutionizes the customer-company relationship and CX. It enables companies to provide efficient, personalized, and responsive support, ultimately leading to customer retention, increased customer value, and a more positive brand image. As organizations continue to leverage the capabilities of LLMs, the landscape of customer service is evolving into a more dynamic and customer-centric ecosystem.

This paper demonstrates a comparison between various model performances and evaluates them on the basis of the quality of response generated. We have compared GOOGLE/FLAN-T5-XXL with GOOGLE/FLAN-T5-BASE, and GOOGLE/FLAN-T5-SMALL and observed that the XXL model outperforms the other LLMs in the provided task. Each model is posed with the same questions.

In the future, Sahaay can access PDFs, Videos, Audio, and other files to extract relevant information about, for example, student activities, research work and innovation carried out by BVM. This multimodal capability has the potential to change forever the way we interact with websites and retrieve information in much less time.

## ACKNOWLEDGMENT

The authors would like to express their deepest appreciation to the research facility provided at TRS BVM Laboratory for encouraging multi-disciplinary collaborative research within the campus. We'd also like to thank Birla Vishvakarma Mahavidyalaya (Engineering College) for allowing us to experiment with the innovation on their website.